\documentclass[letterpaper, 10 pt, conference]{ieeeconf}  

\IEEEoverridecommandlockouts     
\usepackage{cite}
\usepackage{amsmath,amssymb,amsfonts}
\usepackage{algorithmic}
\usepackage{graphicx}

\usepackage{textcomp}
\usepackage{xcolor}
\usepackage{hyperref}
\hypersetup{
  colorlinks=true,  
  linkcolor=black,   
  citecolor=green,   
  urlcolor=magenta  
}

\title{\LARGE \bf
Click to Grasp: Zero-Shot Precise\\ Manipulation via Visual Diffusion Descriptors
}

\author{Nikolaos Tsagkas$^{1,2}$, Jack Rome$^{1}$, Subramanian Ramamoorthy$^{1,2}$, Oisin Mac Aodha$^{1}$, Chris Xiaoxuan Lu$^{3}$ 
\thanks{This work was supported by the United Kingdom Research and Innovation (grant EP/S023208/1), EPSRC Centre for Doctoral Training in Robotics and Autonomous Systems (RAS).}
\thanks{$^{1}$School of Informatics, University of Edinburgh, UK\newline{\tt\small\{n.tsagkas, j.a.rome, s.ramamoorthy, oisin.macaodha\}@ed.ac.uk}}%
\thanks{$^{2}$Edinburgh Centre for Robotics, UK.}%
\thanks{$^{3}$Department of Computer Science, University College London, UK. 
        {\tt\small xiaoxuan.lu@ucl.ac.uk}}%
}

\begin{document}

\maketitle
\thispagestyle{empty}
\pagestyle{empty}

\begin{abstract}
Precise manipulation that is generalizable across scenes and objects remains a persistent challenge in robotics. Current approaches for this task heavily depend on having a significant number of training instances to handle objects with pronounced visual and/or geometric part ambiguities. Our work explores the grounding of fine-grained part descriptors for precise manipulation in a zero-shot setting by utilizing web-trained text-to-image diffusion-based generative models. We tackle the problem by framing it as a dense semantic part correspondence task. Our model returns a gripper pose for manipulating a specific part, using as reference a user-defined click from a source image of a visually different instance of the same object. We require no manual grasping demonstrations as we leverage the intrinsic object geometry and features. Practical experiments in a real-world tabletop scenario validate the efficacy of our approach, demonstrating its potential for advancing semantic-aware robotics manipulation.\newline Web page:~\texttt{\url{https://tsagkas.github.io/click2grasp}} 

\end{abstract}

\section{INTRODUCTION}


Developing semantic-aware manipulation models, with the ability to generalize across scenes and objects, remains an open research challenge in robotics. 
While recent approaches have predominantly concentrated on object-level manipulation, emphasizing the use of language as a communication interface between users and robots, little evidence supports the efficacy of existing methods in achieving precise part-level manipulation. 
This gap is particularly evident in settings containing pronounced visual and/or geometric ambiguity, such as distinguishing between the left and right arm of a stuffed toy or discerning the front-left from the back-right leg of a chair. This paper delves into a core aspect of this challenge: \textit{zero-shot grounding of fine-grained part descriptors for precise manipulation.} 

\begin{figure}[t!]
  \centering 
\includegraphics[width=\columnwidth]{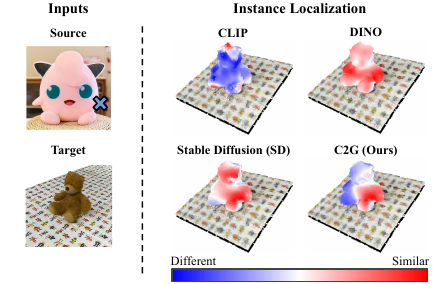}
  \caption{Zero-shot localization heatmaps for identifying the \textit{left arm} of the target stuffed toy from a single click demonstration from the top left source image. \textbf{CLIP} leads to higher activations in irrelevant areas when querying with a natural language prompt, i.e., ``left arm''. \textbf{DINO} achieves high similarity scores for both arm instances, without a clear preference for the left arm. \textbf{SD} features lead to high activations on the left side, but are not localized on the arm. Our \textbf{C2G} approach correctly identifies the left arm.}
  \label{fig:teaser_img}
\end{figure}

An attractive way of framing this problem involves defining the desired interaction area on a specific object category within a source image, thus enabling subsequent reference across various target instances of that object class in the world. 
Visual cues offer distinct advantages over verbal instructions as language may either lack the granularity to describe precise interaction areas (Fig.~\ref{fig:teaser_img}) or when communication barriers exist~\cite{miller2023unknown}. 

The ability to generalize between diverse source and target instances is important for enabling robots to perform a wide variety of manipulation related tasks from limited instruction~\cite{wang2023d3fields}.

One prevalent approach for tackling related tasks in a zero-shot manner is the fusion of features from web-trained foundation models into scene representations, such as point clouds and neural fields~\cite{firoozi2023foundation}. This methodology has been applied successfully in various tasks, from tabletop object manipulation~\cite{shen2023F3RM,wang2023d3fields} to semantic mapping~\cite{tsagkas2023vlfields,shafiullah2023clipfields,huang2023visual,conceptgraphs}, showcasing its versatility and effectiveness. 
Despite these achievements, a notable limitation persists: the granularity of the derived features often does not extend beyond the object-level, as opposed to more local parts (see Fig.~\ref{fig:teaser_img}). 

To address this, some methods have adopted the approach of utilizing extracted activations from these models as priors~\cite{sundaresan2023kite, shridhar2021cliport}, thereby anchoring fine-grained semantic features more effectively. 
However, this necessitates a considerable number of demonstrations, rendering the method non-zero-shot. 
Conversely, fully end-to-end frameworks~\cite{rt12022arxiv, brohan2023rt2} do not completely leverage established well-understood methodologies (e.g., inverse kinematics, scene representations, etc.~), which makes these models notably data-intensive, and their decision making process opaque. 
Even so, the development of extensive robotics datasets has only now started~\cite{embodimentcollaboration2023open, rt12022arxiv,brohan2023rt2, shafiullah2023dobbe}, and scaling trajectory collection, for diverse scenes, object categories, and tasks, to the level of current large image or text datasets remains challenging.

Recently, diffusion models have become the prevailing choice for text-to-image generation~\cite{sd, ho2022classifierfree, nichol2022glide, ramesh2022hierarchical, saharia2022photorealistic}. 
Similar to prior vision models~\cite{caron2021emerging}, there has been growing interest in utilizing the intermediate features as representations for downstream vision tasks~\cite{couairon2022diffedit}. In particular, even in the zero-shot setting, these feature maps have led to state-of-the-art performance in the task of dense semantic correspondence~\cite{zhang2023tale,luo2023dhf,tang2023emergent}.

In addressing the aforementioned manipulation challenge, and drawing inspiration from progress in generative diffusion models, we frame the precise part manipulation task as a dense semantic correspondence problem. 
More specifically, we develop a multi-view consistent scene representation from a multi-camera setup. 
This representation is then integrated with features derived from a diffusion models' noising process~\cite{sd}, making use of the intermediate diffusion activations. 
Using a visually different source image as reference, we define an interaction location in 2D and localize it in the target 3D scene, disambiguating between similar instances of the part type of interest by identifying self-similar part instances.
We validate the efficacy of our method with experiments in a real-world tabletop scenario with different object classes. 
We summarise our contributions as follows: 
\begin{enumerate}
\item We identify the problem of precise part manipulation in the presence of visually and geometrically ambiguous object parts in a generalizable way.

\item We present \textit{Click to Grasp} (C2G), a modular method that takes calibrated RGB-D images of a tabletop and user-defined part instances in diverse source images as input, and produces gripper poses for interaction. Our pipeline effectively disambiguates between visually similar but semantically different concepts (e.g., left vs right arms).

\item We outline an optimization-based approach for solving for gripper poses that eliminates the need for manual demonstrations by solely leveraging the intrinsic geometry and features of a scene of interest.

\end{enumerate}

\begin{figure*}
  \centering
    \includegraphics[width=\textwidth]{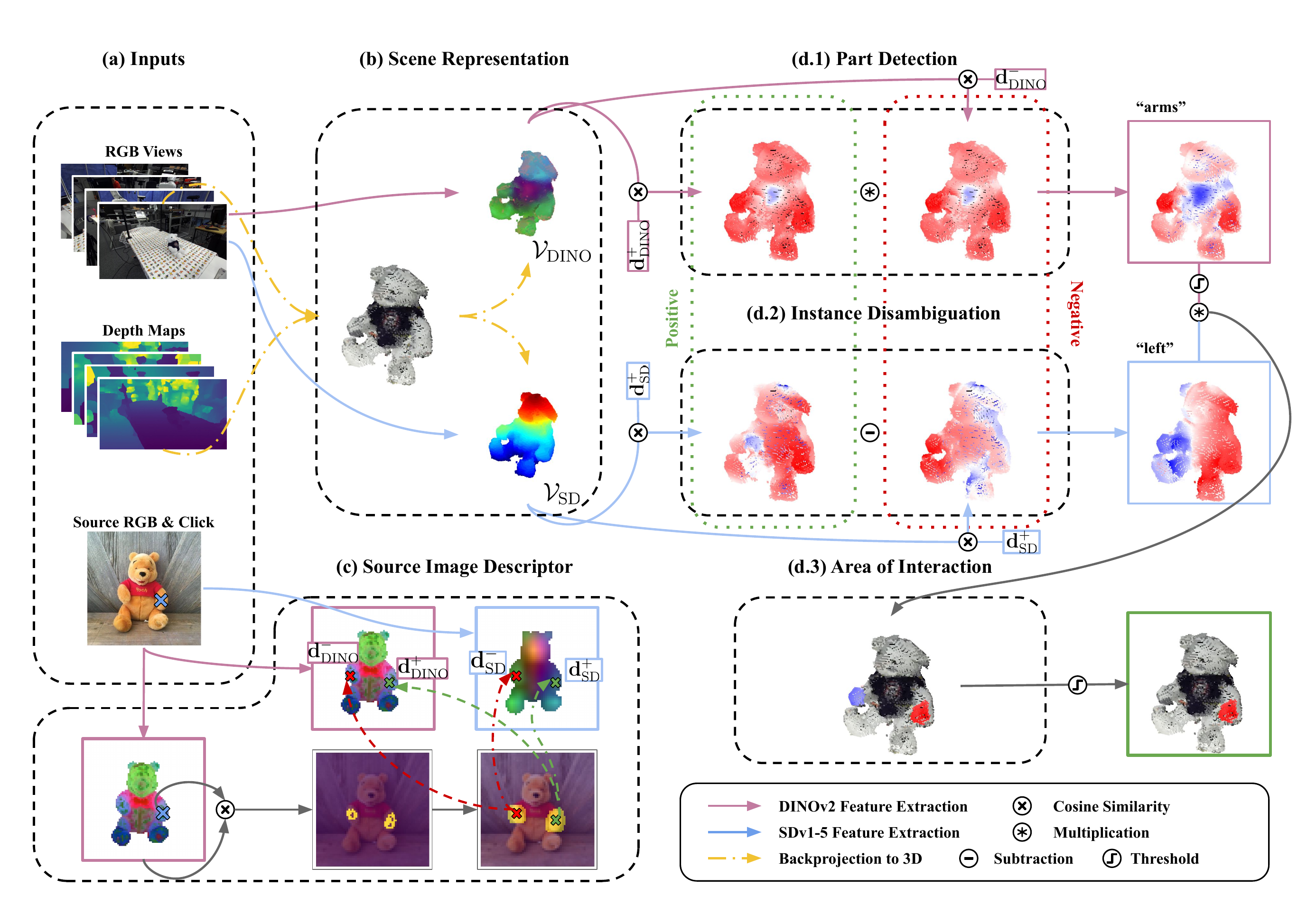}
\caption{\textbf{Perception Modules} $\mathcal{O}_{\mathcal{D}}$, $\mathcal{O}_{\mathcal{G}}$: (a) RGB-D images of the tabletop scene and a random source image of the object class are used as input, along with a single user-defined click, indicating the interaction area (Section~\ref{ssec:problem_formulation}). (b) Images are lifted in 3D space by back-projection and interpolation of RGB values, densities, and features from DINO and SD (Section~\ref{ssec:scene_representation}). (c) DINO and SD features from the source image are extracted and used to localize instances of the user-defined part, automatically identifying them as positive (same instance type) or negative (different instance type), and extracting visual descriptors (Section~\ref{ssec:descriptors}). (d) DINO descriptors localize corresponding parts in the 3D scene, while SD descriptors disambiguate between instances, resulting in a 3D mask identifying the proposed interaction area (Section~\ref{ssec:area_of_interaction}).}
  \label{fig:perception_pipeline}
\end{figure*}

\section{RELATED WORK}

\subsection{Fine-Grained Semantic Object Manipulation}
Precise object manipulation is a key component for a variety of applications. A category of recent works addresses this problem mainly by incorporating language as a natural interface between humans and robots, leveraging the latest advancements from the field of natural language processing. One recent approach uses Large Language Models (LLMs) as high-level planners~\cite{liang2023code, huang2023voxposer} for reasoning via generated code. However, the lack of connection between such LLMs and grounded visual perception means that performance is upper bounded by the vision model's performance. For example, interacting with the front-right leg of a chair requires accurate detection of not only the different legs, but understanding the concept of the front versus the back of the chair. 
Another approach is to use fully end-to-end models to map visual observations and language instructions to a discrete action space (i.e., vision-language action models)~\cite{rt12022arxiv,brohan2023rt2}. However, such models require hundreds of thousands of trajectories from real-world demonstrations for  training, and do not to fully leverage well understood components such as motion planning, etc. An alternative solution utilizes strong priors from vision foundations models to learn specific skills~\cite{shridhar2021cliport, sundaresan2023kite} and jointly encodes them with language instructions. However, a large number of task-specific demonstrations is again required. 
We argue that it is not necessarily always straightforward to use language to describe a precise area of interaction on an object. Instead, visual annotations can be more descriptive of the desired area of interaction. Furthermore, we posit that, in the context of simple grasping tasks, well-established robotics methodologies for manipulation are effective when compared to deploying opaque, end-to-end, approaches that can be cumbersome to train and deploy.

\subsection{Dense Visual Descriptors for Manipulation}
Learning robust, dense visual descriptors, that are adaptable to varying viewing conditions and are applicable across diverse instances of object classes has long been integral to the task of manipulation~\cite{sundaresan2020learning,ganapathi2020learning,manuelli2021keypoints,florence2019selfsupervised,pmlr-v87-florence18a, nerf-supervision}. 
However, the rise of large, web-trained foundation models, has highlighted the efficacy of representations derived from self-supervised models like DINO~\cite{caron2021emerging,oquab2023dinov2} or vision-language models like CLIP~\cite{radford2021learning}. The quality of these features can be enhanced by fusing and combining them from multiple images of a scene into spatial representations. In F3RM~\cite{shen2023F3RM}, DINO or CLIP features are distilled into a neural field, with 3D descriptors obtained via demonstrations by querying an implicit function in 3D coordinates sampled at the location of interaction. However, this scheme necessitates dense views and entails time-consuming re-training when the scene changes. In D$^3$Fields~\cite{wang2023d3fields}, a dynamic representation is introduced, which leverages four camera views and four different foundation models for masking objects~\cite{liu2023grounding, kirillov2023segany, cheng2022xmem} and extracting and tracking descriptors~\cite{oquab2023dinov2}. The latter is utilized for learning object dynamics and planning manipulation tasks with model-predictive control. In our experiments we find that the features extracted by the models used in the aforementioned prior work cannot consistently disambiguate between different object part instances (e.g., telling left part instances from right ones as in Fig.~\ref{fig:teaser_img}).  

\subsection{Diffusion Derived Descriptors}
Text-to-image diffusion models~\cite{sd, ho2022classifierfree, nichol2022glide, ramesh2022hierarchical, saharia2022photorealistic} have recently been shown to be effective at tasks beyond image generation. Particularly Stable Diffusion (SD)~\cite{sd}, which leverages a text-conditioned Latent Diffusion Model, has become widely adopted as a large pre-trained foundation model for diverse downstream computer vision tasks. Example tasks include zero-shot classification~\cite{li2023diffusion, clark2023texttoimage}, segmentation~\cite{xu2022odise, khani2023slime}, and image editing~\cite{couairon2022diffedit, goel2023pair, kawar2023imagic, Tumanyan_2023_CVPR}. Most relevant to our use case is the utilization of intermediate U-Net features from SD as visual descriptors for the task of 2D dense semantic correspondence~\cite{hedlin2023unsupervised,zhang2023tale,luo2023dhf,tang2023emergent}. We highlight two key insights from these approaches. First, although all aforementioned SD-based methods rely on the denoising (generation) process, \cite{luo2023dhf} demonstrated that the noising (inversion) process actually retains crucial semantic visual information, which is otherwise lost. Second, evidence of the complementary nature of SD and DINO features was presented in~\cite{zhang2023tale}. 
In our work, we integrate findings from~\cite{zhang2023tale} and~\cite{luo2023dhf}, by combining feature maps from the inverse process of SD with those from DINO, to create visual descriptors that are capable of resolving part instances. 
Nonetheless, a notable constraint persists in that correspondence between SD features can fail in cases where the pose of the target and source objects are very different~\cite{zhang2024telling}.
We confirm and acknowledge this limitation in our experiments in the 3D world. 
Finally, concurrent to our work, there have been recent attempts from the vision literature to resolve part instance confusion in the context of semantic correspondence estimation~\cite{zhang2024telling,mariotti2024improving}. 
However, unlike those works we do not require any additional training steps or make strong assumptions about the intrinsic 3D geometry of the objects we interact with.


\section{C2G: CLICK TO GRASP}

\subsection{Problem Formulation}
\label{ssec:problem_formulation}
We consider a tabletop scene consisting of any instance of a specific object class. In our experiments, we investigate single-object scenes. However, our method can be easily modified to work with multiple objects, by simply integrating vision-language features, for selecting a specific object to manipulate, based on its visual attributes (e.g., ``the brown teddy bear''). We assume the scene is observable from $N$ distinct cameras, each with known extrinsic parameters, providing RGB images ${I}_t\in\mathbb{R}^{N\times H\times W \times 3}$ and depth maps ${D}_t\in\mathbb{R}^{N\times H\times W}$, which we utilize to develop a 3D scene representation $\mathcal{F}$, as described in Section~\ref{ssec:scene_representation}. Furthermore, our setup incorporates a source RGB image ${I}_s\in\mathbb{R}^{H\times W \times 3}$, showcasing a \emph{different} instance from the same object class. This source instance may exhibit significant visual and geometric differences from those present on the target tabletop scene. An additional input includes a user-specified 2D coordinate $\mathbf{x}_s = (u,v)$ in the pixel space of the source image ${I}_s$, pinpointing a precise part for interaction (e.g., the left arm of a stuffed toy) in two-dimensional space. 

\begin{figure*}[t]
  \centering
  \includegraphics[width=\textwidth]{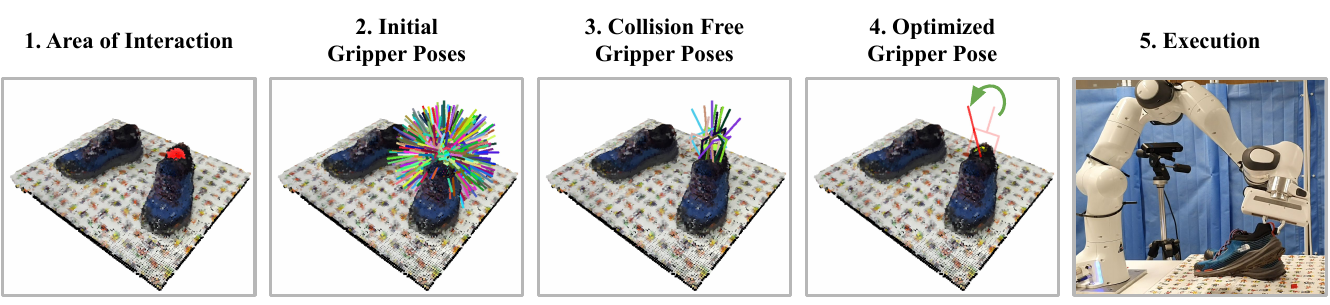}
  \caption{\textbf{Manipulation module} $\mathcal{O}_\mathcal{T}$: Given the proposed area of interaction $\mathcal{A}$, random gripper poses are initialised at each coordinate $\textbf{x}\in\mathcal{A}$. Then, collision-free poses are retained and optimized, using Eq.~\ref{eq:optimizer}. The gripper pose with the lowest final loss score is then sent to the motion planner.}
  \label{fig:manipulation_module}
\end{figure*}

Our goal is to create a model $\mathcal{O}$ that, for any given target scene and source image, enables the precise determination of the transformation $\textbf{T}\in\mathbb{SE}(3)$ to be applied to a robotic arm's end effector. This transformation facilitates the manipulation of an object in the target scene, aligning it with the most visually and semantically similar part specified by the user in the source image. Specifically, our model $\textbf{T} = \mathcal{O}({I}_t, {D}_t | {I}_s, (u, v))$ is structured into three main components, $\mathcal{O} = (\mathcal{O}_\mathcal{D}, \mathcal{O}_\mathcal{G}, \mathcal{O}_\mathcal{T})$. 
$\mathcal{O}_\mathcal{D}$ processes the source image to produce a descriptor $\textbf{d}$, capturing the visual and semantic characteristics of the user-selected point, formalized as $\textbf{d} = \mathcal{O}_\mathcal{D}({I}_s, (u, v))$, which is analyzed in Section~\ref{ssec:descriptors}. A grounding module $\mathcal{O}_\mathcal{G}$ addresses the challenge of matching $\textbf{d}$ with the target scene, identifying a set of coordinates as a 3D \textit{area of interaction} $\mathcal{A}$, represented as $\mathcal{A} = \mathcal{O}_\mathcal{G}(\textbf{d}, {I}_t, {D}_t)$, as described in Section~\ref{ssec:area_of_interaction}. Finally, the manipulation module $\mathcal{O}_\mathcal{T}$ refines the rotational and translational elements of $\textbf{T}$, targeting the vicinity of $\mathcal{A}$. 
This enables the robot to adeptly engage with the designated object segment, denoted as $\textbf{T} = \mathcal{O}_\mathcal{T}(\mathcal{F}, \mathcal{A}, \textbf{d})$. Illustrations of the perception modules $\mathcal{O}_\mathcal{D}, \mathcal{O}_\mathcal{G}$ are provided in Fig.~\ref{fig:perception_pipeline}, while the manipulation module $\mathcal{O}_\mathcal{T}$ is detailed in Fig.~\ref{fig:manipulation_module}.

\subsection{Scene Representation}
\label{ssec:scene_representation}
To represent the target scene we use the implicit descriptor fields representation from \cite{wang2023d3fields}. 
This enables the fusing of RGB-D observations and their corresponding feature maps into a single, multi-view consistent, and differentiable 3D representation $\mathcal{F}(\textbf{x})$, where $\textbf{x}\in\mathbb{R}$ is any 3D point in space. 
We preserve the signed distance to the surface $\textbf{s} \in \mathbb{R}$ as an output of $\mathcal{F}$, and modify it to incorporate two additional implicit field functions that return not only multi-view consistent features $\textbf{f}_{\text{DINO}} \in \mathcal{R}^{N_{\text{DINO}}}$ that have been extracted from DINO, but also $\textbf{f}_{\text{SD}} \in \mathcal{R}^{N_{\text{SD}}}$ from SD: 

\begin{equation}
    \textbf{s} = \mathcal{F}_{\text{s}}(\textbf{x}), \textbf{f}_{\text{SD}} = \mathcal{F}_{\text{f}}^{\text{SD}}(\textbf{x}),  \textbf{f}_{\text{DINO}} = \mathcal{F}_{\text{f}}^{\text{DINO}}(\textbf{x}).
\end{equation}

We extract features from each of the target ${I}_t$ and the source ${I}_s$ RGB images. For DINO, we keep the feature map from the final layer, following prior arts~\cite{zhang2023tale,shen2023F3RM,wang2023d3fields}. For SD, we choose to deploy the inversion process, by noisying the input images and extracting the intermediate feature maps from the UNet, similarly to~\cite{luo2023dhf}. However, we only keep the output from the first timestep and the $4^{th}$ layer. This choice allows us to maintain a low dimensional representation, which is descriptive enough for our task, as it has been visually demonstrated that the feature maps from the middle layers of the U-Net in SD provide a good balance between lower level concepts, such as shapes and outlines, and higher level ones, like texture~\cite{luo2023dhf,zhang2023tale,tang2023emergent,hedlin2023unsupervised}.

\subsection{Source Image Descriptor}\label{ssec:descriptors}
The user-specified coordinates $(u,v)$ in the pixel space of the source image indicate the specific part of the object class of interest that is targeted for manipulation in the tabletop scene. 
We assume that the source image can be any instance of the object class, with the only condition that all the interaction areas of interest are visible. For example, both arms of the stuffed toy need to be visible to interact with the right or the left one. We then extract the relevant feature maps from the source image using both DINO and SD, and denote them as  $\textbf{F}_{\text{DINO}}\in\mathbb{R}^{H'\times W' \times N_{\text{DINO}}}$, $ \textbf{F}_{\text{SD}}\in\mathbb{R}^{H''\times W'' \times N_{\text{SD}}}$. 

Many object categories of interest can contain repeated parts, e.g.,~a stuffed toy can have both a left and right arm. 
These repeated part instances pose a challenge when attempting to find a user specified location from the source image in the target scene. 
To address this, we first rely upon DINO's localization capabilities to identify all instances of the user defined part in the \emph{source} image. 
We compute the cosine similarity between the normalized $\textbf{F}_{\text{DINO}}$ and $\textbf{F}_{\text{DINO}}(u,v)$, and after performing min-max normalization we select the top decile of values, as visualized in Fig.~\ref{fig:perception_pipeline}~(c). We empirically find that this selection step is adequate for roughly identifying all possible interaction instances. 
However, further processing is required to avoid noisy similarity masks in the source image. 
Dilation followed by erosion proves sufficient to rectify this issue, resulting in the formation of convex hulls. The centroids of which lie inside all visible areas of interaction and can provide reliable descriptors. 
While this work concentrates on the binary distinction between left and right, the methodology we propose can be adapted to address more complex scenarios involving objects with many instances of the same part, such as the legs of a chair or the handles of different drawers.


We extract descriptors $\textbf{d}_{\text{DINO}}$ and $\textbf{d}_{\text{SD}}$ from  $\textbf{F}_{\text{DINO}}$ and $\textbf{F}_{\text{SD}}$ for the coordinates of the identified centroids and assign the spatially closest to the user defined click as \textit{positive} ($\textbf{d}_{\text{DINO}}^+$, $\textbf{d}_{\text{SD}}^+$) and the other(s) as \textit{negative} ($\textbf{d}_{\text{DINO}}^-$, $\textbf{d}_{\text{SD}}^-$). We empirically find that identifying both the \textit{positive} and \textit{negative} samples is imperative, both for accurately performing part detection in the scene, but most importantly, for solving semantics ambiguities, as is explained next.

\subsection{Identifying Target Scene Area of Interaction}
\label{ssec:area_of_interaction}
The target scene representation and extracted source image positive and negative descriptors are utilized for proposing a 3D area of interaction $\mathcal{A}$. Given the implicit nature of the scene representation, we first convert it into a voxel grid $\mathcal{V}$, spanning a volume $H_t \times W_t \times L_t$ over the tabletop, with voxel size equal to $\delta$. After initializing the grid, we query $\mathcal{F}$ and assign to each voxel a $\textbf{f}_{\text{DINO}}$ and a $\textbf{f}_{\text{SD}}$ feature embedding. From this, we define two voxel feature grids: $\mathcal{V}_{\text{DINO}}$ and $\mathcal{V}_{\text{SD}}$.\newline

\noindent\textbf{Part detection}: First, $\textbf{d}_{\text{DINO}}^+$ and $\textbf{d}_{\text{DINO}}^-$ are utilized for grounding all instances of the user-defined part in the 3D scene. We compute cosine similarity between $\mathcal{V}_{\text{DINO}}$ and the two descriptors. As shown in Fig.~\ref{fig:perception_pipeline}~(d-1), both the $\textbf{d}_{\text{DINO}}^+$ and $\textbf{d}_{\text{DINO}}^-$ lead to higher activations in all areas of the present part instances. Given the minuscule disparity between two similarities, DINO fails to localize a clear candidate area of interaction. Consequently, we utilize these features for part detection, increasing the similarity scores in the relevant areas, and decreasing it everywhere else, by multiplying the similarity scores from the two descriptors:

\begin{equation}
    \text{SIM}_{\text{DINO}} = \frac{\mathbf{d}_{\text{DINO}}^+ \cdot \mathcal{V}_{\text{DINO}}}{\|\mathbf{d}_{\text{DINO}}^+\| \cdot \|\mathcal{V}_{\text{DINO}}\|} \cdot  
    \frac{\mathbf{d}_{\text{DINO}}^- \cdot \mathcal{V}_{\text{DINO}}}{\|\mathbf{d}_{\text{DINO}}^-\| \cdot \|\mathcal{V}_{\text{DINO}}\|}
\end{equation}

\noindent Finally, we threshold the min-max normalized $\text{SIM}_{\text{DINO}}$ scores at $\theta_{\text{DINO}}$ to identify possible areas of interaction. \newline

\noindent\textbf{Part Instance Disambiguation}: We then rely upon the SD features, for the purpose of selecting the most relevant part instance, from the ones detected with DINO. Again, we calculate the cosine similarity between the SD descriptors and $\mathcal{V}_{\text{SD}}$, which lead to relatively noisy similarity heatmaps. As illustrated in Fig. \ref{fig:perception_pipeline}~(d-2), the similarity scores are generally higher at the broader region of the corresponding user-defined instance, which are however not reliable enough for instance selection, as relatively high scores exist in all parts. We find that by adding a contrastive component, computing the difference between the similarity scores from $\textbf{d}_{\text{SD}}^+$ and $\textbf{d}_{\text{SD}}^-$, respectively, the resulting heatmap becomes more suitable for instance disambiguation:

\begin{equation}
    \text{SIM}_{\text{SD}} = \frac{\mathbf{d}_{\text{SD}}^+ \cdot \mathcal{V}_{\text{SD}}}{\|\mathbf{d}_{\text{SD}}^+\| \cdot \|\mathcal{V}_{\text{SD}}\|} - 
    \frac{\mathbf{d}_{\text{SD}}^- \cdot \mathcal{V}_{\text{SD}}}{\|\mathbf{d}_{\text{SD}}^-\| \cdot \|\mathcal{V}_{\text{SD}}\|}
\end{equation}

\noindent\textbf{Area of Interaction}: Finally, we utilize the resulting similarity scores $\text{SIM}_{\text{DINO}}$, $\text{SIM}_{\text{SD}}$ to propose an interaction area $\mathcal{A}$ in the scene. We multiply per coordinate the min-max normalized similarity scores of the thresholded $\text{SIM}_{\text{DINO}}$ and the min-max normalized $\text{SIM}_{\text{DINO}}$ similarity:

\begin{equation}
\mathcal{A}' = \max(\text{SIM}_{\text{DINO}}, \theta_{\text{DINO}}) \cdot \text{SIM}_{\text{SD}}.
\end{equation} 

\noindent The voxels of $\mathcal{A}'$ that correspond to zero similarity scores are discarded and the values of the remaining voxels are min-max normalized. We threshold the final scores, keeping the voxels that correspond to the top quartile. This set of coordinates is the proposed area of interaction $\mathcal{A}$, thus grounding the user defined click from the source image into the 3D space of the target scene.

\subsection{Gripper Pose Optimization}
\label{ssec:gripper}
For the purpose of interacting with an object in the target scene, we must solve for the transformation $\textbf{T} \in \mathbb{SE}(3)$ to be applied on the robot's end effector to determine its final 6-DoF pose in the world coordinate system. 
Here, $\mathbb{SE}(3) = \{ (\textbf{R}, \textbf{t}) \,|\, \textbf{R} \in \mathbb{SO}(3),\, \textbf{t} \in \mathbb{R}^3 \}$,  with rotation matrix $\textbf{R}$ and translation vector $\textbf{t}$. Our approach is similar to~\cite{shen2023F3RM}. 
However, we do not require any manipulation demonstrations to optimize the pose. Instead, we rely on the gripper's and scene's geometry to identify a collision free solution. 


\begin{figure*}[ht]
  \centering
  \includegraphics[width=\textwidth]{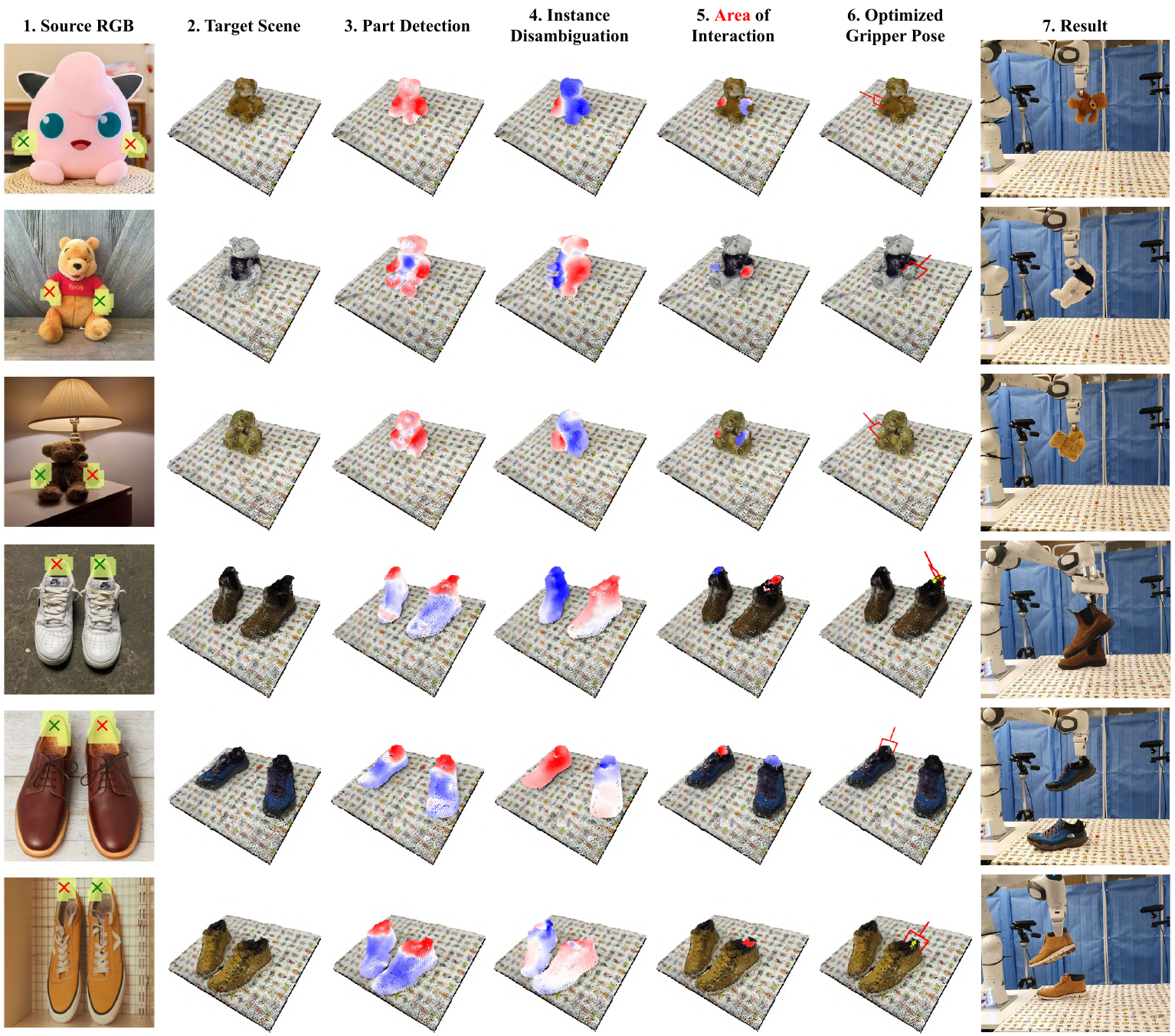}
  \caption{\textbf{Visualization of results} for three stuffed toys and three shoes manipulation experiments: (1) Source images along with the detected $(u,v)$ coordinates of the \textit{positive} and \textit{negative} part instances. (2) Reconstructed 3D scene. (3) Part similarity heatmap. (4) Part instance disambiguation. (5) Identifying the \textcolor{red}{area of interaction} versus the most similar \textcolor{blue}{part instance}. (6) Optimized gripper pose. (7) Real-world object manipulation.}
  \label{fig:descriptor_extraction}
\end{figure*}

Given the area of interaction $\mathcal{A}$, we initialize transforms $\{\textbf{T}\}=\{(\textbf{R}, \textbf{t})\}$ for the robot's end-effector, by setting as $\textbf{t}$ the coordinates of the selected voxels and use random rotation matrices. The center point between the gripper fingers is then utilized as the mean $\mu$ of an elongated 3D Gaussian distribution for sampling a set of $M$ points $\mathcal{X}_i$, that acts as a sampler of the implicit function at the predicted area of interaction. We find that elongating the distribution orthogonally to the axis of movement of the gripper fingers assists our optimizer to align better with the object's geometry. We also generate another point cloud $\mathcal{X}_g$, which corresponds to the gripper's geometry, for the purpose of detecting collisions with the scene. We then leverage the differentiable nature of the implicit function $\mathcal{F}$ to compute gradients for our optimizer when querying with the transformed points of $\mathcal{X}_i$ and $\mathcal{X}_g$. 

Our gripper loss function consists of three components: 
\begin{equation}
  \label{eq:optimizer}
  \begin{aligned}
  \mathcal{L}(\textbf{T}; \mathbf{d}_{\text{DINO}}^+) =
  &-\frac{1}{M_s} \sum_{\textbf{x}_i\in\mathbf{T}\mathcal{X}_i} \overbrace{\lambda_{\text{a}}^f \mathcal{L}_a^f(\textbf{x}_i, \mathbf{d}_{\text{DINO}}^+) + \lambda_{\text{a}}^g\mathcal{L}_a^g(\textbf{X}_i)}^{\text{Attractive force}}\\ 
  &-\frac{1}{M_g} \sum_{ \textbf{x}_g\in\mathbf{T}\mathcal{X}_g} \underbrace{\lambda_{\text{r}} \mathcal{L}_r(\textbf{x}_g)}_{\text{Repulsive force}} + \underbrace{\lambda_{\text{reg}}  \mathcal{L}_{\text{reg}}(\textbf{T})}_{\text{Regularization term}}.
  \end{aligned}
\end{equation}

\noindent First, we include two components that act as attraction forces. $\mathcal{L}_a^f(\textbf{x}_i, \mathbf{d}_{\text{DINO}}^+)$ pulls the $\textbf{T}\mathcal{X}_i$ points closer to areas that maximize the cosine similarity between the DINO scene features $\textbf{f}_{\text{DINO}}$ and the descriptor $\mathbf{d}_{\text{DINO}}^+$, while $\mathcal{L}_a^g$ pulls the same points close to the surface of the scene, leveraging the signed distance output of $\mathcal{F}$. We found that only DINO features are enough for our optimizer for not diverging the gripper from the proposed area of interaction. Second, we include a repulsive force guided by the gripper geometry $\textbf{T}\mathcal{X}_g$, pushing its point cloud away from the scene, so that collisions are avoided. 
The constituent losses are defined as: 

\begin{equation}
   \mathcal{L}_a^f(\textbf{x}, \textbf{d}) = \frac{\mathcal{F}_f(\textbf{x}) \cdot \textbf{d}}{\|\mathcal{F}_f(\textbf{x})\| \cdot \|\textbf{d}\|}, 
    \mathcal{L}_r(\textbf{x}) = -\mathcal{L}_a^g(\textbf{x}) = \mathcal{F}_s(\textbf{x})  
\end{equation}


The equilibrium of these forces is a collision-free gripper pose that can interact successfully with the target 3D identified location. We also add a regularizer term: 
\begin{equation}
    \mathcal{L}_{\text{reg}}(\textbf{T}) = \|\text{\textbf{R}}\|_2 + \|\text{\textbf{t}}\|_2,
\end{equation}
\noindent which encourages the values of the rotation and translation matrices to remain low, thus further stabilizing the optimization. We repeat this optimization for all collision free poses 
and select the one with the lowest final cost.

\section{EXPERIMENTS}

\subsection{Experiment Setup}
We setup a tabletop scene, which is observed by four ZED2i stereo cameras, the extrinsics of which were calibrated in the world frame of a Franka Panda arm. 
We use DINOv2 ViT-B/14 (which we refer to as DINO) and Stable Diffusion v1-5 (SD) as our visual feature extractors. 
We constrain the target object's pose so that it does not vary greatly from the object in the source image for at least two camera views from which we extract the SD features. 
This choice is important as feature extraction in SD is not robust to large global transformations~\cite{zhang2024telling}.

We focus our experiments on successfully manipulating stuffed toys and shoes. 
These items are good candidates for evaluation because they contain visual and geometric symmetries which necessitates the need for our C2G approach but at the same time can be manipulated in a tabletop setting. 
Traditional used object categories in semantic correspondence evaluation (e.g., vehicles, animals, or furniture) are not valid for grasping experiments.  

A simple UI was also developed to streamline the interaction.  
Once a target object is positioned on the tabletop, the user is prompted with a corresponding source image of the same object class.
We used three distinct object instances for both the shoes and the stuffed toys class, and three web source images for each. 
We assessed the ability of a model to generate successful gripper poses for each part instance within the respective classes. 
In Section~\ref{ssec:qual} we conducted comparisons against SD and DINO alone, evaluating the accuracy of the proposed gripper poses \emph{offline} using pre-recorded data. 
Each model underwent 36 gripper pose prediction experiments, resulting in a total of 108 evaluations. 
In Section~\ref{ssec:quant} we replicated the 36 C2G experiments in a real-world setting using a real robot arm to inspect its ability to correctly pick an object.



\subsection{Instance Localization Evaluation}
\label{ssec:qual}
First, we compare the ability of C2G to generate plausible gripper poses for grasping user-defined part instance against using only DINO or SD features offline. 
For both $\mathcal{V}_{\text{DINO}}$ and $\mathcal{V}_{\text{SD}}$, we select as area of interaction $\mathcal{A}$ the set of points that has a cosine similarity score with the corresponding descriptors $\textbf{d}_{\text{DINO}}^+$, $\textbf{d}_{\text{SD}}^+$ above $0.85$, respectively. For all three cases, we feed the proposed $\mathcal{A}$ to the optimizer. A pose proposal is marked as successful if the correct instance part lies between the gripper fingers after the pose optimization. 
Results are summarized in Table \ref{tab:qual}.

We observe that C2G is the most accurate in generating plausible poses for interacting with the user-defined part instance. 
Using only DINO features in most cases led to a gripper pose that would grasp the object from the correct part (i.e., the arm of a stuffed toy or the lip of a shoe's opening), however, there was no clear preference towards the correct part instance (i.e., left or right), which is why the success rate is close to $50\%$. 
SD features were always successful in finding a pose in which the gripper would interact with approximately the correct side of the object. 
However, in many cases the gripper would not interact with the precise part of the object.
We found that this issue was most noticeable in the stuffed toys, with the optimized pose leading to interactions in the leg and ear areas. We observed a significant drop in performance when using the source image with the pink stuffed toy, where $5/6$ times the gripper pose was closest to the ear of the toy in the scene, 
This indicates that the generalization capabilities of SD features are poorer.
Nevertheless, SD features led to surprisingly better performance in the case of the shoe class, which we attribute to the fact that the two instances are spatially separated by empty space. 
In comparison, our C2G generated gripper poses were correctly able to identify both the correct part \emph{and} part instance, failing only once.


\subsection{Grasping Evaluation}
\label{ssec:quant}
Finally we implement C2G in a real-world scenario to assess its capability to effectively grasp objects from the intended part instance. 
After optimizing the gripper's pose within the proposed interaction area, we utilized inverse kinematics to map out a manipulation trajectory, commencing from the neutral position of the robot arm. 
A successful grasp is defined as one where the correct part instance is picked up, and the grasp is robust enough such that the object does not fall when lifted. 
We summarize our results in Table \ref{tab:quant}. 
C2G failed to perform a successful grasp in only three out of the 36 scenarios. These failures are attributed to various factors: one resulted from slipping, another from an inaccurate proposal of the interaction area, and the third from inadequate pose prediction.


\begin{table}[t]
\caption{Offline Instance localization results}
\begin{center}
\begin{tabular}{c|cc|c}
    \textbf{Model} & \textbf{Stuffed Toys} $\uparrow$  & \textbf{Shoes} $\uparrow$& \textbf{Total} $\uparrow$ \\ 
    \hline
    \hline
    DINO        & 6/18  & 9/18 & 15/36\\
    SD          & 8/18  & 15/18 & 23/36\\
    C2G (ours)  & \underline{17/18} & \underline{18/18} & \underline{35/36}\\
    \hline
\end{tabular}
\label{tab:qual}
\end{center}
\vspace{-5pt}
\end{table}

\begin{table}[t]
\caption{Real-World Grasping results}
\begin{center}
\begin{tabular}{c|cc|c}

         & \textbf{Stuffed Toys} $\uparrow$ & \textbf{Shoes} $\uparrow$ & \textbf{Per Instance} $\uparrow$\\
    \hline
    \hline
    Left  & 9/9          & 8/9  &  17/18 \\ 
    Right & 8/9          & 8/9  &  16/18 \\
    \hline
    Per Object & 17/18 & 15/18 &   33/36 \\
    \hline
\end{tabular}
\label{tab:quant}
\end{center}
\vspace{-10pt}
\end{table}


\section{CONCLUSIONS}
We presented C2G, a precise manipulation approach for objects exhibiting semantic ambiguities. 
C2G uses a user-defined annotation in the form of a single click from a source image to generate gripper poses in the real-world. 
Our approach builds on prior works that fuse features from web-trained foundation models. 
However, C2G distinguishes among similar instances of parts within tabletop scenarios. 
This is achieved by recognizing part correspondences within a source image of the same object category as the target scene and performing disambiguation of the part instances.
We demonstrated that simply fusing features from foundation models is insufficient for precise grasping in real-world tabletop scenes, whereas our C2G approach led to $92\%$ grasping success.
We identify two limitations of our approach. 
First,  the target object's pose should not vary greatly from the one depicted in the source image, an issue that has been documented in concurrent work~\cite{zhang2024telling}. Second, we currently \textit{do not} utilize the language integration capabilities of SD. 
We acknowledge both of the above limitations as future research directions. 
Our C2G approach is potentially well suited for industrial environments since a single annotation from a generic 2D source image is sufficient to precisely manipulate diverse target object instances of the same class. 







\section*{ACKNOWLEDGMENT}
We would like to thank Peter D.~Fagan and Elle Miller for the valuable discussions. 




\bibliographystyle{IEEEtran}
\bibliography{IEEEabrv,root.bib}

\begin{thebibliography}{10}
\providecommand{\url}[1]{#1}
\csname url@samestyle\endcsname
\providecommand{\newblock}{\relax}
\providecommand{\bibinfo}[2]{#2}
\providecommand{\BIBentrySTDinterwordspacing}{\spaceskip=0pt\relax}
\providecommand{\BIBentryALTinterwordstretchfactor}{4}
\providecommand{\BIBentryALTinterwordspacing}{\spaceskip=\fontdimen2\font plus
\BIBentryALTinterwordstretchfactor\fontdimen3\font minus \fontdimen4\font\relax}
\providecommand{\BIBforeignlanguage}[2]{{%
\expandafter\ifx\csname l@#1\endcsname\relax
\typeout{** WARNING: IEEEtran.bst: No hyphenation pattern has been}%
\typeout{** loaded for the language `#1'. Using the pattern for}%
\typeout{** the default language instead.}%
\else
\language=\csname l@#1\endcsname
\fi
#2}}
\providecommand{\BIBdecl}{\relax}
\BIBdecl

\bibitem{miller2023unknown}
E.~Miller, M.~Durner, M.~Humt, G.~Quere, W.~Boerdijk, A.~M. Sundaram, F.~Stulp, and J.~Vogel, ``Unknown object grasping for assistive robotics,'' in \emph{ICRA}, 2024.

\bibitem{wang2023d3fields}
Y.~Wang, Z.~Li, M.~Zhang, K.~Driggs-Campbell, J.~Wu, L.~Fei-Fei, and Y.~Li, ``D$^3$fields: Dynamic 3d descriptor fields for zero-shot generalizable robotic manipulation,'' \emph{arXiv:2309.16118}, 2023.

\bibitem{firoozi2023foundation}
R.~Firoozi, J.~Tucker, S.~Tian, A.~Majumdar, J.~Sun, W.~Liu, Y.~Zhu, S.~Song, A.~Kapoor, K.~Hausman \emph{et~al.}, ``Foundation models in robotics: Applications, challenges, and the future,'' \emph{arXiv:2312.07843}, 2023.

\bibitem{shen2023F3RM}
W.~Shen, G.~Yang, A.~Yu, J.~Wong, L.~P. Kaelbling, and P.~Isola, ``Distilled feature fields enable few-shot language-guided manipulation,'' in \emph{CoRL}, 2023.

\bibitem{tsagkas2023vlfields}
N.~Tsagkas, O.~M. Aodha, and C.~X. Lu, ``Vl-fields: Towards language-grounded neural implicit spatial representations,'' in \emph{ICRA Workshops}, 2023.

\bibitem{shafiullah2023clipfields}
N.~M.~M. Shafiullah, C.~Paxton, L.~Pinto, S.~Chintala, and A.~Szlam, ``Clip-fields: Weakly supervised semantic fields for robotic memory,'' in \emph{RSS}, 2023.

\bibitem{huang2023visual}
C.~Huang, O.~Mees, A.~Zeng, and W.~Burgard, ``Visual language maps for robot navigation,'' in \emph{ICRA}, 2023.

\bibitem{conceptgraphs}
Q.~Gu, A.~Kuwajerwala, S.~Morin, K.~Jatavallabhula, B.~Sen, A.~Agarwal, C.~Rivera, W.~Paul, K.~Ellis, R.~Chellappa, C.~Gan, C.~{de Melo}, J.~Tenenbaum, A.~Torralba, F.~Shkurti, and L.~Paull, ``Conceptgraphs: Open-vocabulary 3d scene graphs for perception and planning,'' \emph{arXiv:2309.16650}, 2023.

\bibitem{sundaresan2023kite}
P.~Sundaresan, S.~Belkhale, D.~Sadigh, and J.~Bohg, ``Kite: Keypoint-conditioned policies for semantic manipulation,'' \emph{arXiv:2306.16605}, 2023.

\bibitem{shridhar2021cliport}
M.~Shridhar, L.~Manuelli, and D.~Fox, ``Cliport: What and where pathways for robotic manipulation,'' in \emph{CoRL}, 2021.

\bibitem{rt12022arxiv}
A.~Brohan, N.~Brown, J.~Carbajal, Y.~Chebotar, J.~Dabis, C.~Finn \emph{et~al.}, ``Rt-1: Robotics transformer for real-world control at scale,'' \emph{arXiv:2212.06817}, 2022.

\bibitem{brohan2023rt2}
A.~Brohan, N.~Brown, J.~Carbajal, Y.~Chebotar, X.~Chen, K.~Choromanski \emph{et~al.}, ``Rt-2: Vision-language-action models transfer web knowledge to robotic control,'' \emph{arXiv:2307.15818}, 2023.

\bibitem{embodimentcollaboration2023open}
A.~Padalkar, A.~Pooley, A.~Jain, A.~Bewley, A.~Herzog, A.~Irpan, A.~Khazatsky, A.~Rai, A.~Singh, A.~Brohan \emph{et~al.}, ``Open x-embodiment: Robotic learning datasets and rt-x models,'' \emph{arXiv:2310.08864}, 2023.

\bibitem{shafiullah2023dobbe}
N.~M.~M. Shafiullah, A.~Rai, H.~Etukuru, Y.~Liu, I.~Misra, S.~Chintala, and L.~Pinto, ``On bringing robots home,'' \emph{arXiv:2311.16098}, 2023.

\bibitem{sd}
R.~Rombach, A.~Blattmann, D.~Lorenz, P.~Esser, and B.~Ommer, ``High-resolution image synthesis with latent diffusion models,'' in \emph{CVPR}, 2022.

\bibitem{ho2022classifierfree}
J.~Ho and T.~Salimans, ``Classifier-free diffusion guidance,'' \emph{arXiv:2207.12598}, 2022.

\bibitem{nichol2022glide}
A.~Nichol, P.~Dhariwal, A.~Ramesh, P.~Shyam, P.~Mishkin, B.~McGrew, I.~Sutskever, and M.~Chen, ``Glide: Towards photorealistic image generation and editing with text-guided diffusion models,'' in \emph{ICML}, 2022.

\bibitem{ramesh2022hierarchical}
A.~Ramesh, P.~Dhariwal, A.~Nichol, C.~Chu, and M.~Chen, ``Hierarchical text-conditional image generation with clip latents,'' \emph{arXiv:2204.06125}, 2022.

\bibitem{saharia2022photorealistic}
C.~Saharia, W.~Chan, S.~Saxena, L.~Li, J.~Whang, E.~L. Denton, K.~Ghasemipour, R.~Gontijo~Lopes, B.~Karagol~Ayan, T.~Salimans \emph{et~al.}, ``Photorealistic text-to-image diffusion models with deep language understanding,'' in \emph{NeurIPS}, 2022.

\bibitem{caron2021emerging}
M.~Caron, H.~Touvron, I.~Misra, H.~J\'egou, J.~Mairal, P.~Bojanowski, and A.~Joulin, ``Emerging properties in self-supervised vision transformers,'' in \emph{ICCV}, 2021.

\bibitem{couairon2022diffedit}
G.~Couairon, J.~Verbeek, H.~Schwenk, and M.~Cord, ``Diffedit: Diffusion-based semantic image editing with mask guidance,'' in \emph{ICLR}, 2023.

\bibitem{zhang2023tale}
J.~Zhang, C.~Herrmann, J.~Hur, L.~P. Cabrera, V.~Jampani, D.~Sun, and M.-H. Yang, ``A tale of two features: Stable diffusion complements dino for zero-shot semantic correspondence,'' in \emph{NeurIPS}, 2023.

\bibitem{luo2023dhf}
G.~Luo, L.~Dunlap, D.~H. Park, A.~Holynski, and T.~Darrell, ``Diffusion hyperfeatures: Searching through time and space for semantic correspondence,'' in \emph{NeurIPS}, 2023.

\bibitem{tang2023emergent}
L.~Tang, M.~Jia, Q.~Wang, C.~P. Phoo, and B.~Hariharan, ``Emergent correspondence from image diffusion,'' in \emph{NeurIPS}, 2023.

\bibitem{liang2023code}
J.~Liang, W.~Huang, F.~Xia, P.~Xu, K.~Hausman, B.~Ichter, P.~Florence, and A.~Zeng, ``Code as policies: Language model programs for embodied control,'' in \emph{ICRA}, 2023.

\bibitem{huang2023voxposer}
W.~Huang, C.~Wang, R.~Zhang, Y.~Li, J.~Wu, and L.~Fei-Fei, ``Voxposer: Composable 3d value maps for robotic manipulation with language models,'' in \emph{CoRL}, 2023.

\bibitem{sundaresan2020learning}
P.~Sundaresan, J.~Grannen, B.~Thananjeyan, A.~Balakrishna, M.~Laskey, K.~Stone, J.~E. Gonzalez, and K.~Goldberg, ``Learning rope manipulation policies using dense object descriptors trained on synthetic depth data,'' in \emph{ICRA}, 2020.

\bibitem{ganapathi2020learning}
A.~Ganapathi, P.~Sundaresan, B.~Thananjeyan, A.~Balakrishna, D.~Seita, J.~Grannen, M.~Hwang, R.~Hoque, J.~E. Gonzalez, N.~Jamali \emph{et~al.}, ``Learning dense visual correspondences in simulation to smooth and fold real fabrics,'' in \emph{ICRA}, 2021.

\bibitem{manuelli2021keypoints}
L.~Manuelli, Y.~Li, P.~Florence, and R.~Tedrake, ``Keypoints into the future: Self-supervised correspondence in model-based reinforcement learning,'' in \emph{CoRL}, 2021.

\bibitem{florence2019selfsupervised}
P.~Florence, L.~Manuelli, and R.~Tedrake, ``Self-supervised correspondence in visuomotor policy learning,'' in \emph{RA-L}, 2019.

\bibitem{pmlr-v87-florence18a}
P.~R. Florence, L.~Manuelli, and R.~Tedrake, ``Dense object nets: Learning dense visual object descriptors by and for robotic manipulation,'' in \emph{CoRL}, 2018.

\bibitem{nerf-supervision}
L.~Yen-Chen, P.~Florence, J.~T. Barron, T.-Y. Lin, A.~Rodriguez, and P.~Isola, ``Nerf-supervision: Learning dense object descriptors from neural radiance fields,'' in \emph{ICRA}, 2022.

\bibitem{oquab2023dinov2}
M.~Oquab, T.~Darcet, T.~Moutakanni, H.~V. Vo, M.~Szafraniec, V.~Khalidov, P.~Fernandez, D.~Haziza, F.~Massa, A.~El-Nouby, R.~Howes \emph{et~al.}, ``Dinov2: Learning robust visual features without supervision,'' \emph{TMLR}, 2024.

\bibitem{radford2021learning}
A.~Radford, J.~W. Kim, C.~Hallacy, A.~Ramesh, G.~Goh, S.~Agarwal, G.~Sastry, A.~Askell, P.~Mishkin, J.~Clark \emph{et~al.}, ``Learning transferable visual models from natural language supervision,'' in \emph{ICML}, 2021.

\bibitem{liu2023grounding}
S.~Liu, Z.~Zeng, T.~Ren, F.~Li, H.~Zhang, J.~Yang, C.~Li, J.~Yang, H.~Su, J.~Zhu \emph{et~al.}, ``Grounding dino: Marrying dino with grounded pre-training for open-set object detection,'' \emph{arXiv:2303.05499}, 2023.

\bibitem{kirillov2023segany}
A.~Kirillov, E.~Mintun, N.~Ravi, H.~Mao, C.~Rolland, L.~Gustafson, T.~Xiao, S.~Whitehead, A.~C. Berg, W.-Y. Lo \emph{et~al.}, ``Segment anything,'' in \emph{ICCV}, 2023.

\bibitem{cheng2022xmem}
H.~K. Cheng and A.~G. Schwing, ``{XMem}: Long-term video object segmentation with an atkinson-shiffrin memory model,'' in \emph{ECCV}, 2022.

\bibitem{li2023diffusion}
A.~C. Li, M.~Prabhudesai, S.~Duggal, E.~Brown, and D.~Pathak, ``Your diffusion model is secretly a zero-shot classifier,'' in \emph{ICCV}, 2023.

\bibitem{clark2023texttoimage}
K.~Clark and P.~Jaini, ``Text-to-image diffusion models are zero shot classifiers,'' in \emph{NeurIPS}, 2023.

\bibitem{xu2022odise}
J.~Xu, S.~Liu, A.~Vahdat, W.~Byeon, X.~Wang, and S.~De~Mello, ``Open-vocabulary panoptic segmentation with text-to-image diffusion models,'' in \emph{CVPR}, 2023.

\bibitem{khani2023slime}
A.~Khani, S.~A. Taghanaki, A.~Sanghi, A.~M. Amiri, and G.~Hamarneh, ``Slime: Segment like me,'' in \emph{ICLR}, 2024.

\bibitem{goel2023pair}
V.~Goel, E.~Peruzzo, Y.~Jiang, D.~Xu, N.~Sebe, T.~Darrell, Z.~Wang, and H.~Shi, ``Pair-diffusion: Object-level image editing with structure-and-appearance paired diffusion models,'' \emph{arXiv:2303.17546}, 2023.

\bibitem{kawar2023imagic}
B.~Kawar, S.~Zada, O.~Lang, O.~Tov, H.~Chang, T.~Dekel, I.~Mosseri, and M.~Irani, ``Imagic: Text-based real image editing with diffusion models,'' in \emph{CVPR}, 2023.

\bibitem{Tumanyan_2023_CVPR}
N.~Tumanyan, M.~Geyer, S.~Bagon, and T.~Dekel, ``Plug-and-play diffusion features for text-driven image-to-image translation,'' in \emph{CVPR}, 2023.

\bibitem{hedlin2023unsupervised}
E.~Hedlin, G.~Sharma, S.~Mahajan, H.~Isack, A.~Kar, A.~Tagliasacchi, and K.~M. Yi, ``Unsupervised semantic correspondence using stable diffusion,'' in \emph{NeurIPS}, 2023.

\bibitem{zhang2024telling}
J.~Zhang, C.~Herrmann, J.~Hur, E.~Chen, V.~Jampani, D.~Sun, and M.-H. Yang, ``Telling left from right: Identifying geometry-aware semantic correspondence,'' in \emph{CVPR}, 2024.

\bibitem{mariotti2024improving}
O.~Mariotti, O.~Mac~Aodha, and H.~Bilen, ``Improving semantic correspondence with viewpoint-guided spherical maps,'' in \emph{CVPR}, 2024.

\end{thebibliography}

\end{document}